\documentclass[11pt]{article}
\usepackage[preprint]{acl}
\usepackage{times,latexsym}
\usepackage[T1]{fontenc}
\usepackage[utf8]{inputenc}
\usepackage{microtype,inconsolata}
\usepackage{graphicx,amsmath,amssymb,booktabs,multirow,xcolor}
\usepackage{colortbl,tabularx}
\usepackage{placeins,float}
\usepackage{fontawesome5}
\definecolor{rolaqblue}{HTML}{326AA5}
\colorlet{rolaqshade}{rolaqblue!8}

\newcommand{\method}{\textsc{RoLaQ}}
\newcommand{\tablemethod}{\textbf{\method{}}}
\newcommand{\R}{\mathbb{R}}
\newcommand{\E}{\mathbb{E}}
\newcommand{\std}[1]{{\scriptsize\,$\pm$#1}}
\newif\ifdraft\drafttrue

\newcommand{\nCPMMinutes}{34}

\newcommand{\nCPMNowinGainBest}{1.1}

\newcommand{\nCPMNowinOurs}{59.4}

\newcommand{\nCPMTKone}{0.85}

\newcommand{\nDSAvgBestBase}{45.3}

\newcommand{\nDSMinutes}{22}

\newcommand{\nDSNowinGainBest}{3.8}
\newcommand{\nDSNowinGsmOSCAR}{18.3}
\newcommand{\nDSNowinGsmOurs}{32.1}
\newcommand{\nDSNowinOurs}{46.7}

\newcommand{\nDSRulerMultiKeyOSCAR}{7.0}
\newcommand{\nDSRulerMultiKeyOurs}{55.0}

\newcommand{\nDSTKone}{0.91}

\newcommand{\nKappaOfficial}{1.041}

\newcommand{\nMLNowinGainBest}{2.8}

\newcommand{\nMLNowinOurs}{60.1}

\newcommand{\nMLTKone}{0.89}

\newcommand{\abDSFullAvg}{48.1}

\newcommand{\abMLFullAvg}{62.6}

\newcommand{\nCaseLosses}{69}
\newcommand{\nCaseTotal}{1319}
\newcommand{\nCaseWins}{277}

\title{RoLaQ: Joint Rotation and Range Calibration\\
for 2-Bit KV Caches in Multi-Head Latent Attention}
\author{%
Shiyu Ji$^{1}$ \quad Peng Ling$^{2}$ \quad
Chen Lin$^{3}$ \quad Shupeng Li$^{3}$ \\
\bfseries Yixuan Wang$^{1}$ \quad Qingfu Zhu$^{1}$ \quad
Wanxiang Che$^{1}$\thanks{Corresponding author.} \\
{\normalfont $^1$Research Center for Social Computing and Interactive Robotics,} \\
{\normalfont Harbin Institute of Technology, China} \\
{\normalfont $^2$Tsinghua University, China} \\
{\normalfont $^3$Baidu Inc., China} \\
{\normalfont\texttt{\{syji,car\}@ir.hit.edu.cn}} \\
\faIcon{github-alt}~\url{https://github.com/noforit/RoLaQ}
}

\begin{document}
% The second title-footnote symbol is dagger; maketitle resets the counter.
\setcounter{footnote}{1}
\maketitle
\begin{abstract}
Multi-head latent attention (MLA) compresses the content of a key--value
cache into a shared latent per token, making low-bit storage attractive for
long-context inference.
At two bits, errors in this representation perturb both attention
matching and value readout, with effects that depend on error direction
and reconstruction scale.
We introduce \method{} (Rotation-calibrated Latent Quantization),
an offline method for joint rotation and range calibration.
\method{} refines a shared rotation within subspaces defined by the
frozen key and value projections, and jointly learns quantization ranges
whose thresholds and reconstruction gains are coupled to control
the shared signal's response.
Prediction matching and attention-output supervision guide calibration
with model weights frozen.
Across three MLA models and six benchmarks, \method{} achieves the highest
average accuracy among the compared 2-bit methods both without and with
token protection.
On DeepSeek-V2-Lite, an SGLang implementation provides $3.96\times$
the BF16 KV capacity under the same memory budget and $2.51\times$
its decode throughput with 256K input tokens.
\end{abstract}
\section{Introduction}

\begin{figure}[t]
\centering
\includegraphics[width=\columnwidth]{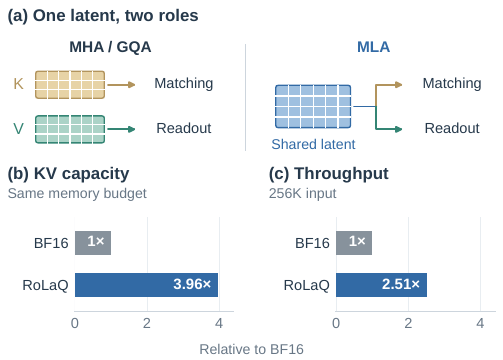}
\caption{(a) MLA shares a content cache between keys and values.
(b,c) On one A100 80GB, \method{} gives DeepSeek-V2-Lite
$3.96\times$ the BF16 capacity at equal memory budgets and
$2.51\times$ its decode throughput with four concurrent requests
and 256K inputs.}
\label{fig:teaser}
\end{figure}

Long contexts and concurrent requests increase the memory and bandwidth
costs of the key--value (KV) cache during autoregressive decoding
\citep{transformer,kivi}.
Multi-head latent attention (MLA), introduced in DeepSeek-V2, reduces
these costs by caching a compact content latent shared across heads
\citep{deepseekv2}.
MLA and its variants now underpin models including DeepSeek-V3,
GLM-5.3, and Hy4 \citep{deepseekv3,glm53,hy4}.
Even with this architectural compression, the cache continues to grow
with context length and serving concurrency.
KV cache quantization offers further savings by lowering the storage
precision of cached states \citep{kivi,kvquant,gear}.
We focus on quantizing MLA's content latent to two bits while preserving
its two roles in attention matching and value readout.

KIVI exploits distinct K/V statistics with per-channel key and
per-token value quantization \citep{kivi}.
KVQuant and GEAR further control distortion through outlier handling
and residual error compensation \citep{kvquant,gear}.
Rotations make low-bit representations easier to quantize by
redistributing outliers \citep{quarot,spinquant,rotatekv}.
Attention-aware calibration further adapts these transformations to
the computations that consume the cache, as in OptR's attention-output
reconstruction objective \citep{optr}.
These methods primarily target multi-head or grouped-query attention
with separate key and value caches.

In MLA, every head derives both content keys and values from the
same quantized latent through distinct projections
(Figure~\ref{fig:teaser}a).
A single perturbation can therefore change both the matching scores
that select information and the values aggregated under those scores.
OSCAR provides covariance-aware rotations for this shared cache
\citep{oscar}.
These projection paths guide our joint calibration of the latent basis
and quantization range.

Two observations guide our design (\S\ref{sec:structure}).
Joint K/V projection energy is unevenly distributed across latent
directions, so equal-magnitude errors can produce different distortions.
Reconstruction gain also affects prediction fidelity even when the
rotation is fixed.
Because rotation changes the quantizer's input distribution,
we learn its boundaries alongside the latent basis.
Together, these observations motivate joint rotation and range learning
with a controlled reconstruction response.

We introduce \method{}, which jointly learns a shared rotation within
K/V-defined subspaces and a quantization threshold per layer,
with model weights frozen.
Each threshold determines its reconstruction gain, allowing the
quantization assignments to adapt while controlling the shared
signal's reconstruction response.
A BF16 teacher supplies next-token distributions and attention
outputs: prediction KL aligns the complete model, while attention-output
supervision directly guides the individual layers.
Calibration takes roughly 30 minutes on one A100 GPU.
At inference, \method{} uses a packed 2-bit content cache and absorbs
read-side rotations into MLA's projections, eliminating standalone
rotation multiplications when attending to the cached history.

We evaluate \method{} on DeepSeek-V2-Lite, Moonlight-16B-A3B, and
MiniCPM3-4B across six benchmarks.
It achieves the highest average accuracy among the compared 2-bit
methods both with and without token protection.
Without protection, it improves on the strongest baseline by
\nCPMNowinGainBest{} to \nDSNowinGainBest{} points
(Table~\ref{tab:main}).
Retaining selected tokens in BF16 further improves accuracy.
RULER retrieval at 8K and 16K shows that calibration gains extend
beyond the 4K training context.
Figure~\ref{fig:teaser}b,c shows the SGLang gains for DeepSeek-V2-Lite:
$3.96\times$ the BF16 cache capacity at equal memory budgets and
$2.51\times$ its decode throughput with 256K input on one A100 80GB.

Our contributions are as follows:
\begin{itemize}
% Keep each contribution together; this changes only break opportunities.
\interlinepenalty=10000
\item We develop joint rotation and range calibration for MLA,
using a compact K/V search space and coupled thresholds and gains
to control the shared latent's response.
\item We demonstrate accuracy gains across three MLA models alongside
long-context retrieval improvements, with ablations explaining the
roles of rotation geometry, quantizer response, and calibration
objectives.
\item We implement packed INT2 serving in SGLang with fused cache
writes and optimized attention paths, delivering greater KV capacity,
higher long-context throughput, and lower first-token tail latency.
\end{itemize}

\section{MLA Cache Quantization: Structure and Challenges}
\label{sec:structure}

\subsection{One Cache, Two Roles}
For each token $t$, MLA caches a content latent $c_t\in\R^d$ and a
decoupled RoPE key $k^R_t\in\R^{d_r}$ \citep{deepseekv2}.
Each head $h$ obtains its content key and value through frozen
up-projections, $k_{t,h}=c_tW_{K,h}^{\top}$ and
$v_{t,h}=c_tW_{V,h}^{\top}$.
Using row vectors, a query with content part $q_h$ and positional part
$q^R_h$ produces
\begin{equation}
\begin{aligned}
\alpha_{t,h}&=\operatorname{softmax}_{t}\!\left[
\tau\big(q_h k_{t,h}^{\top}+q^R_h{k^R_t}^{\top}\big)\right],\\
y&=\textstyle\sum_h\sum_t\alpha_{t,h}v_{t,h}W_{O,h},
\end{aligned}
\label{eq:mla}
\end{equation}
where $\tau$ is the attention scale, softmax runs over causally visible
positions, and $W_{O,h}$ is head $h$'s output-projection block.
The shared latent therefore determines both where attention looks
through the keys and what it reads through the values.
Its quantization error propagates through both roles.

We quantize $c_t$ to two bits and retain the lower-dimensional RoPE key $k^R_t$
in BF16 because of its quantization sensitivity \citep{snapmla}.
An orthogonal rotation $R$ spreads outliers before quantization
\citep{quarot,spinquant}.
Writing $Q$ for quantization and $\operatorname{DQ}$ for dequantization,
every head uses $\hat c_t=\operatorname{DQ}(Q(c_tR))R^{\top}$
for matching and readout.

\subsection{Projection-Dependent Distortion}
Let $e_t=\hat c_t-c_t$ be the reconstruction error.
Its effects on the attention logit and value are
\begin{equation}
\Delta s_{t,h}=\tau q_hW_{K,h}e_t^{\top},\qquad
\Delta v_{t,h}=e_tW_{V,h}^{\top}.
\label{eq:cache-error}
\end{equation}
Errors of equal magnitude can affect the two projections differently
depending on their direction.
Let $W_K$ and $W_V$ stack the corresponding up-projections across heads.
For $X\in\{K,V\}$, the squared projection error is
$\|eW_X^{\top}\|^2=eW_X^{\top}W_Xe^{\top}$.
We define
\begin{equation}
H_X=\frac{W_X^{\top}W_X}{\operatorname{tr}(W_X^{\top}W_X)},
\qquad H=H_K+H_V.
\label{eq:kv-geometry}
\end{equation}
Trace normalization gives both projections equal total weight.
The quadratic form $eHe^{\top}$ sums normalized squared K/V errors.
Each eigenvalue of $H$ measures this error along its unit eigenvector.

\begin{figure}[t]
\centering
\includegraphics[width=\columnwidth]{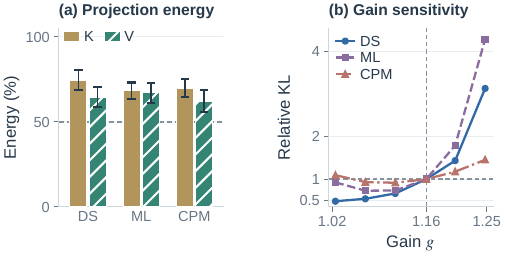}
\caption{Two observations on shared-latent quantization.
(a) K/V energy in the leading half of the joint spectrum
(mean $\pm$ layer std.; dashed: 50\%).
(b) Prediction KL at 8K with OSCAR's rotation and threshold fixed,
normalized by its value at $g_0=1.16$ (vertical dashed line).
DS, ML, and CPM denote DeepSeek-V2-Lite, Moonlight, and MiniCPM3.}
\label{fig:motivation}
\end{figure}

Figure~\ref{fig:motivation}a shows that the leading half of $H$'s
eigenvectors captures 62--75\% of each projection's energy across the
three models, averaged over layers.
This common basis separates directions by their influence on both
projections, guiding the shared rotation search.

\subsection{Sensitivity to Reconstruction Scale}
Beyond error direction, reconstruction scale affects both matching
and readout.
The threshold controls four-level assignments, while the gain scales
the reconstructed signal shared by both paths.

Figure~\ref{fig:motivation}b isolates this effect by fixing OSCAR's
rotation and threshold and varying only its gain.
Values below one indicate lower KL than the baseline with fixed $g_0=1.16$.
On 8K-token documents, adjusting the gain alone reduces
DeepSeek-V2-Lite's next-token KL to BF16 by 52\% within the sweep.
Together, these observations motivate joint calibration of the shared
rotation and quantization range.
Appendix~\ref{app:observations} details both experiments and the full spectra.

\section{Method}
\label{sec:method}
\begin{figure*}[t]
\centering
\includegraphics[
  width=\textwidth,
  trim=3.65bp 0bp 3.65bp 0bp,
  clip
]{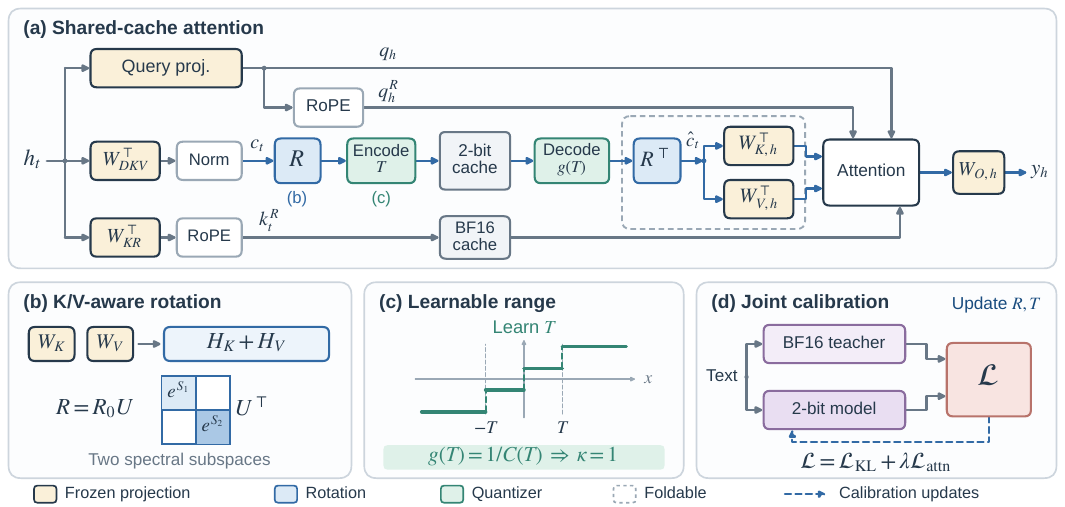}
\caption{Overview of \method{}.
(a) A shared 2-bit content cache supplies both K/V paths.
The dashed box groups the read-side rotation with the projections that absorb it.
(b) $H_K+H_V$ defines two spectral subspaces for learning the shared rotation.
(c) The learned threshold determines the four-level assignments and gain
$g(T)=1/C(T)$.
(d)~A frozen BF16 teacher guides joint calibration through prediction KL
and attention-output loss.}
\label{fig:method}
\end{figure*}

\method{} jointly calibrates the shared representation and its quantizer
to preserve attention matching and value readout (Figure~\ref{fig:method}).
The \emph{K/V-aware shared rotation} uses the joint projection geometry
to organize its search space (\S\ref{sec:rotation}).
The \emph{learnable quantization range} adapts the decision boundary to
this basis while preserving the centered signal's linear response
(\S\ref{sec:range}).
Joint calibration uses prediction matching and attention-output supervision
to learn both components in the frozen model
(Figure~\ref{fig:method}(d); \S\ref{sec:calibration}).

\subsection{K/V-Aware Shared Rotation}
\label{sec:rotation}
We use $H$ from Eq.~\ref{eq:kv-geometry} to divide rotation refinement
between directions with high and low joint projection energy.
Refining within these subspaces preserves this distinction while allowing
the quantization basis to adapt to calibration data.
Both subspaces contribute to the key and value paths.

Starting from OSCAR's covariance-aware rotation $R_0$ \citep{oscar},
we express the geometry in the initial rotated coordinates as
$R_0^{\top}HR_0=U\Lambda U^{\top}$.
We sort the eigenvalues in descending order and split $U=[U_1,U_2]$
into two equal-dimensional subspaces.
Within each, a learnable skew-symmetric generator defines the refinement:
\begin{equation}
\begin{aligned}
R&=R_0U\operatorname{diag}\!\big(e^{S_1},e^{S_2}\big)U^{\top},\\
S_i&=A_i-A_i^{\top},\quad A_i\in\R^{\frac d2\times\frac d2}.
\end{aligned}
\label{eq:structured}
\end{equation}
This parameterization keeps $R$ orthogonal throughout calibration;
zero-initializing $A_1,A_2$ recovers $R_0$.
The two generators use $d^2/2$ trainable parameters, half the number
in a full $d\times d$ generator.
The calibration objective selects the rotation within this structured space.
Section~\ref{sec:geometry-analysis} evaluates this choice of rotation
search space.

\subsection{Response-Preserving Range Calibration}
\label{sec:range}
The range controls both the quantization assignments and the strength
of the reconstructed signal.
We adapt the assignments to the learned rotation while anchoring the
linear response of the shared signal to its full-precision reference.
We standardize each group of $G$ rotated coordinates using its mean $\mu$
and standard deviation $\sigma$ \citep{oscar}.
For a coordinate $z$, reconstruction takes the form
\begin{equation}
x=\frac{z-\mu}{\sigma},\qquad \hat z=\mu+\sigma g\,a_T(x).
\label{eq:quantizer}
\end{equation}
Here $a_T$ assigns $x$ to four ordered levels
$\{\pm a_{\mathrm{in}},\pm a_{\mathrm{out}}\}$ using boundaries $(-T,0,T)$,
with $a_{\mathrm{out}}=3a_{\mathrm{in}}$.
The threshold $T$ controls assignment to the inner or outer levels,
while the gain $g$ scales reconstruction around the group mean.

To separate systematic scale change from residual distortion, we use
$x\sim\mathcal N(0,1)$ as a reference distribution.
The quantized signal decomposes as $g\,a_T(x)=\kappa x+\epsilon$,
where $\E[x\epsilon]=0$ \citep{bussgang1952,bussgang} and
\begin{equation}
\kappa=gC(T),\qquad C(T)=\E[x\,a_T(x)].
\label{eq:response}
\end{equation}
The coefficient $\kappa$ measures the linear response to the input.
In the original group scale,
$\hat z-\mu=\kappa(z-\mu)+\sigma\epsilon$.
Changing $T$ and $g$ therefore changes the response of the centered
signal reused by both attention paths.
Appendix~\ref{app:response} derives $C(T)$.

To enforce unit response under the Gaussian reference, we set the
gain from the threshold:
\begin{equation}
g(T)=\frac{1}{C(T)}\quad\Longrightarrow\quad\kappa=1.
\label{eq:gain}
\end{equation}
Each layer learns $T_\ell$ (Appendix~\ref{app:thresholds}), and its gain
follows from this constraint.
The assignments can therefore adapt while preserving unit response.
Section~\ref{sec:why} tests this response after calibration across context
lengths, complementing the initial gain sweep in Figure~\ref{fig:motivation}b.

% Declare the main table early so it can appear at the top of page 5.
\begin{table*}[t]
\centering
\small
\setlength{\tabcolsep}{4pt}
\begingroup
% Each pair reads nowin / s4r16.
\newcommand{\windowpair}[2]{%
  \makebox[1.8em][r]{#1}\hspace{0.20em}%
  {\color{gray}/}\hspace{0.20em}\makebox[1.8em][r]{#2}}
\begin{tabularx}{\linewidth}{@{}>{\columncolor{white}[0pt][\tabcolsep]}l*{6}{>{\centering\arraybackslash}X}>{\centering\arraybackslash\columncolor{white}[\tabcolsep][0pt]}X@{}}
\toprule
\multirow{2}{*}{\textbf{Method}} & \multicolumn{3}{c}{\textbf{Knowledge \& QA}} & \multicolumn{1}{c}{\textbf{Math}} & \multicolumn{2}{c}{\textbf{Code}} & \multirow{2}{*}{\textbf{Avg.}} \\
\cmidrule(lr){2-4}\cmidrule(lr){5-5}\cmidrule(lr){6-7}
 & MMLU & CMMLU & TriviaQA & GSM8K & HumanEval & MBPP & \\
\midrule
\multicolumn{8}{@{}c@{}}{\textbf{DeepSeek-V2-Lite}} \\
\textcolor{gray}{BF16} & \textcolor{gray}{58.0} & \textcolor{gray}{64.1} & \textcolor{gray}{65.2} & \textcolor{gray}{37.4} & \textcolor{gray}{28.0} & \textcolor{gray}{43.4} & \textcolor{gray}{49.4} \\
RTN & \windowpair{40.5}{44.9} &
  \windowpair{45.8}{51.8} &
  \windowpair{41.5}{59.6} &
  \windowpair{7.0}{15.1} &
  \windowpair{12.2}{18.9} &
  \windowpair{17.6}{28.0} &
  \windowpair{27.4}{36.4} \\
QuaRot & \windowpair{53.2}{53.8} &
  \windowpair{58.6}{60.6} &
  \windowpair{60.4}{63.4} &
  \windowpair{25.5}{30.5} &
  \windowpair{\textbf{25.6}}{25.0} &
  \windowpair{33.8}{38.6} &
  \windowpair{42.9}{45.3} \\
KVarN & \windowpair{49.9}{53.7} &
  \windowpair{56.0}{59.1} &
  \windowpair{57.8}{63.4} &
  \windowpair{23.6}{27.9} &
  \windowpair{20.1}{26.2} &
  \windowpair{32.6}{37.0} &
  \windowpair{40.0}{44.6} \\
OSCAR & \windowpair{50.8}{50.3} &
  \windowpair{57.8}{57.6} &
  \windowpair{60.9}{62.2} &
  \windowpair{18.3}{18.9} &
  \windowpair{22.6}{23.8} &
  \windowpair{36.6}{38.6} &
  \windowpair{41.1}{41.9} \\
\rowcolor{rolaqshade}
\tablemethod{} & \windowpair{\textbf{55.1}}{\textbf{55.7}} &
  \windowpair{\textbf{61.5}}{\textbf{62.1}} &
  \windowpair{\textbf{64.1}}{\textbf{64.9}} &
  \windowpair{\textbf{32.1}}{\textbf{34.2}} &
  \windowpair{25.4}{\textbf{27.0}} &
  \windowpair{\textbf{41.8}}{\textbf{42.5}} &
  \windowpair{\textbf{46.7}}{\textbf{47.7}} \\
\midrule
\multicolumn{8}{@{}c@{}}{\textbf{Moonlight-16B-A3B}} \\
\textcolor{gray}{BF16} & \textcolor{gray}{70.0} & \textcolor{gray}{78.2} & \textcolor{gray}{66.4} & \textcolor{gray}{73.5} & \textcolor{gray}{47.0} & \textcolor{gray}{53.4} & \textcolor{gray}{64.7} \\
RTN & \windowpair{61.2}{65.6} &
  \windowpair{65.1}{72.8} &
  \windowpair{46.7}{64.8} &
  \windowpair{44.1}{61.6} &
  \windowpair{23.8}{41.5} &
  \windowpair{37.4}{46.2} &
  \windowpair{46.4}{58.8} \\
QuaRot & \windowpair{63.1}{65.8} &
  \windowpair{68.9}{73.4} &
  \windowpair{48.1}{64.4} &
  \windowpair{49.7}{67.5} &
  \windowpair{32.3}{41.5} &
  \windowpair{39.0}{\textbf{50.8}} &
  \windowpair{50.2}{60.6} \\
KVarN & \windowpair{45.6}{63.0} &
  \windowpair{45.0}{67.9} &
  \windowpair{17.6}{58.5} &
  \windowpair{21.2}{64.1} &
  \windowpair{12.2}{\textbf{47.0}} &
  \windowpair{16.6}{45.6} &
  \windowpair{26.4}{57.7} \\
OSCAR & \windowpair{66.9}{\textbf{67.9}} &
  \windowpair{73.9}{75.4} &
  \windowpair{58.6}{63.7} &
  \windowpair{64.3}{69.4} &
  \windowpair{34.8}{43.3} &
  \windowpair{45.4}{48.4} &
  \windowpair{57.3}{61.3} \\
\rowcolor{rolaqshade}
\tablemethod{} & \windowpair{\textbf{67.4}}{\textbf{67.9}} &
  \windowpair{\textbf{74.9}}{\textbf{76.1}} &
  \windowpair{\textbf{63.2}}{\textbf{65.3}} &
  \windowpair{\textbf{66.5}}{\textbf{70.5}} &
  \windowpair{\textbf{40.4}}{44.3} &
  \windowpair{\textbf{47.9}}{50.1} &
  \windowpair{\textbf{60.1}}{\textbf{62.3}} \\
\midrule
\multicolumn{8}{@{}c@{}}{\textbf{MiniCPM3-4B}} \\
\textcolor{gray}{BF16} & \textcolor{gray}{64.4} & \textcolor{gray}{72.9} & \textcolor{gray}{36.2} & \textcolor{gray}{75.7} & \textcolor{gray}{74.4} & \textcolor{gray}{51.8} & \textcolor{gray}{62.6} \\
RTN & \windowpair{54.3}{59.3} &
  \windowpair{59.2}{67.8} &
  \windowpair{28.5}{33.5} &
  \windowpair{35.4}{65.0} &
  \windowpair{22.6}{68.3} &
  \windowpair{31.6}{45.6} &
  \windowpair{38.6}{56.6} \\
QuaRot & \windowpair{61.7}{62.5} &
  \windowpair{69.6}{70.1} &
  \windowpair{33.9}{34.9} &
  \windowpair{68.5}{71.4} &
  \windowpair{\textbf{70.1}}{70.1} &
  \windowpair{45.0}{48.4} &
  \windowpair{58.1}{59.6} \\
KVarN & \windowpair{59.3}{62.8} &
  \windowpair{66.7}{71.5} &
  \windowpair{27.1}{34.9} &
  \windowpair{52.4}{72.5} &
  \windowpair{47.6}{70.7} &
  \windowpair{39.4}{\textbf{51.4}} &
  \windowpair{48.7}{60.6} \\
OSCAR & \windowpair{\textbf{62.9}}{63.0} &
  \windowpair{69.9}{\textbf{71.6}} &
  \windowpair{31.9}{34.9} &
  \windowpair{\textbf{71.2}}{74.2} &
  \windowpair{67.1}{70.1} &
  \windowpair{47.0}{49.2} &
  \windowpair{58.3}{60.5} \\
\rowcolor{rolaqshade}
\tablemethod{} & \windowpair{\textbf{62.9}}{\textbf{63.5}} &
  \windowpair{\textbf{71.0}}{71.2} &
  \windowpair{\textbf{34.7}}{\textbf{35.3}} &
  \windowpair{70.5}{\textbf{74.6}} &
  \windowpair{67.1}{\textbf{73.8}} &
  \windowpair{\textbf{50.3}}{49.7} &
  \windowpair{\textbf{59.4}}{\textbf{61.4}} \\
\bottomrule
\end{tabularx}
\endgroup

\caption{Downstream accuracy (\%) with 2-bit content caches.
Each pair reports \emph{w.o.\ / w.\ token protection}
(without / with 4 sink and 16 recent tokens retained in BF16).
BF16 is a shared reference; bold marks the best 2-bit scores
at the displayed precision within each setting.}
\label{tab:main}
\end{table*}

\subsection{Joint Calibration and Inference}
\label{sec:calibration}
The rotation changes the distribution seen by the quantizer, so we learn
the rotation generators and thresholds together across all $L$ layers.
The backbone remains frozen.
For the same input text, the BF16 model supplies target next-token
distributions and attention outputs.
We optimize
\begin{equation}
\begin{aligned}
\mathcal L&=\mathcal L_{\mathrm{KL}}+\lambda\mathcal L_{\mathrm{attn}},\\
\mathcal L_{\mathrm{KL}}&=\frac{1}{|\mathcal I|}\sum_{t\in\mathcal I}
\mathrm{KL}\!\big(p_t^{\mathrm{BF16}}\,\|\,p_t^{\mathrm{2bit}}\big),\\
\mathcal L_{\mathrm{attn}}&=\frac{1}{L}\sum_{\ell=1}^L
\frac{\|Y_\ell^{\mathrm{2bit}}-Y_\ell^{\mathrm{BF16}}\|_F^2}
{\|Y_\ell^{\mathrm{BF16}}\|_F^2}.
\end{aligned}
\label{eq:calibration}
\end{equation}
Here $\mathcal I$ contains the scored token positions, $p_t$ is the
next-token distribution, and $Y_\ell$ collects attention outputs after
$W_O$ at those positions.
Prediction KL aligns the complete quantized model with its BF16 teacher.
The auxiliary output loss, weighted by $\lambda$, supplies a direct
signal to each layer through both attention matching and value readout,
following output-aware calibration \citep{optr}.
Normalizing by teacher output energy balances layers with different scales.

We initialize the rotation and threshold from OSCAR and set the gain
using Eq.~\ref{eq:gain}.
The forward pass uses the hard quantizer in Eq.~\ref{eq:quantizer}.
A straight-through estimator passes gradients to its inputs \citep{ste},
while a smooth threshold surrogate supplies gradients for bin assignment.
The threshold also receives gradients through $g(T_\ell)$.
Calibration settings are in \S\ref{sec:experiments}; quantizer constants
and checkpoint selection are in Appendix~\ref{app:calibration}.

At inference, the learned rotation is exported as one dense matrix per
layer, and each new latent is rotated once and stored as 2-bit codes
with its group statistics.
For $\hat z_t=\operatorname{DQ}(Q(c_tR))$, the read-side rotation in
Figure~\ref{fig:method}(a)'s dashed box folds into MLA's absorbed projections:
\begin{equation}
\begin{aligned}
\widetilde q_h&=q_hW_{K,h}R,\\
\widetilde W_{O,h}&=R^{\top}W_{V,h}^{\top}W_{O,h},\\
y&=\textstyle\sum_h\big(\sum_t\alpha_{t,h}\hat z_t\big)
\widetilde W_{O,h}.
\end{aligned}
\label{eq:absorption}
\end{equation}
The content logit is $\tau\widetilde q_h\hat z_t^{\top}$, so both
matching and readout operate in the rotated basis.
The calibrated rotation, thresholds, and gains remain fixed throughout serving.
Only the group statistics are computed from incoming tokens.

\section{Experiments}
\label{sec:experiments}

\subsection{Setup}
\paragraph{Models and benchmarks.}
We evaluate DeepSeek-V2-Lite \citep{deepseekv2}, Moonlight-16B-A3B
\citep{moonlight}, and MiniCPM3-4B \citep{minicpm3} on six benchmarks:
MMLU and CMMLU for knowledge \citep{mmlu,cmmlu}, TriviaQA for question answering
\citep{triviaqa}, GSM8K for mathematics \citep{gsm8k}, and HumanEval and MBPP
for code generation \citep{humaneval,mbpp}.
We also evaluate long-context retrieval on the eight RULER retrieval tasks
\citep{ruler}.

\paragraph{Baselines and cache settings.}
We compare with round-to-nearest quantization (RTN), QuaRot \citep{quarot},
KVarN \citep{kvarn}, and OSCAR \citep{oscar}.
QuaRot and KVarN are applied to the shared MLA latent; OSCAR uses its official
MLA recipe.
All comparisons use 2-bit content caches and BF16 RoPE keys.
We evaluate every method both without token protection and with the first
4 sink tokens and most recent 16 tokens retained in BF16
\citep{streamingllm,kivi,oscar}.
Accuracy is evaluated with simulated quantization in Hugging Face
Transformers, giving all baselines a common evaluation framework.

\paragraph{Calibration.}
We calibrate \method{} on 1.43M Pile tokens \citep{pile} with the model frozen
and select checkpoints by held-out prediction KL.
Main results average three seeds; component, objective, and
random-subspace ablations use one.
Calibration takes about \nDSMinutes{} minutes for the 16B models and
\nCPMMinutes{} minutes for MiniCPM3 on one A100 GPU.
Appendix~\ref{app:protocol} provides model configurations, evaluation protocols,
baseline implementations, and calibration hyperparameters.

\subsection{Main Results}
\method{} achieves the highest six-task average on all three models under
both protection settings (Table~\ref{tab:main}).
We first examine the setting in which every content token is quantized,
which directly tests how well each method preserves accuracy under 2-bit storage.
\method{} reaches \nDSNowinOurs{}, \nMLNowinOurs{}, and \nCPMNowinOurs{} on
DeepSeek-V2-Lite, Moonlight, and MiniCPM3, exceeding the strongest baseline by
\nDSNowinGainBest{}, \nMLNowinGainBest{}, and \nCPMNowinGainBest{} points.
The larger margins without protection highlight the benefit of calibrating
the shared representation and its quantization range when the entire content
history passes through the quantizer.

On DeepSeek-V2-Lite, GSM8K accuracy without protection improves from OSCAR's
\nDSNowinGsmOSCAR{} to \nDSNowinGsmOurs{}.
Moreover, \method{} without protection reaches a higher average than the
strongest baseline with 20 protected tokens
(\nDSNowinOurs{} vs.\ \nDSAvgBestBase{}).

Retaining 4 sink and 16 recent tokens in BF16 further improves \method{},
which maintains the highest average accuracy on all three models.
The calibrated cache thus benefits from mixed-precision token protection
while retaining its advantage over the baselines.
Section~\ref{sec:windows} examines protection budgets;
Appendix~\ref{app:seeds} reports all seeds and windows.

Appendix~\ref{app:cases} illustrates these differences with arithmetic
and retrieval examples, comparing the answers generated by BF16
and 2-bit models.

\begin{table}[t]
\centering
\fontsize{8}{9.5}\selectfont
\setlength{\tabcolsep}{1.3pt}
\begin{tabularx}{\linewidth}{@{}>{\columncolor{white}[0pt][\tabcolsep]}l*{8}{>{\centering\arraybackslash}X}>{\centering\arraybackslash\columncolor{white}[\tabcolsep][0pt]}X@{}}
\toprule
\textbf{Method} & S1 & S2 & S3 & MK1 & MK2 & MK3 & MV & MQ & \textbf{Avg.} \\
\midrule
\multicolumn{10}{@{}c@{}}{\textbf{DeepSeek-V2-Lite}} \\
\textcolor{gray}{BF16} & \textcolor{gray}{100.0} & \textcolor{gray}{100.0} & \textcolor{gray}{96.0} & \textcolor{gray}{94.0} & \textcolor{gray}{88.0} & \textcolor{gray}{47.0} & \textcolor{gray}{63.5} & \textcolor{gray}{89.0} & \textcolor{gray}{84.7} \\
RTN & 86.0 & 55.0 & 21.0 & 23.0 & 0.0 & 0.0 & 33.0 & 25.2 & 30.4 \\
QuaRot & \textbf{100.0} & 98.0 & 80.0 & 54.0 & 35.0 & 5.0 & \textbf{64.2} & 72.0 & 63.5 \\
KVarN & \textbf{100.0} & 97.0 & 77.0 & 75.0 & \textbf{58.0} & 7.0 & 50.5 & 74.5 & 67.4 \\
OSCAR & \textbf{100.0} & 89.0 & 70.0 & 52.0 & 7.0 & 1.0 & 55.8 & 62.5 & 54.7 \\
\rowcolor{rolaqshade}
\tablemethod{} & \textbf{100.0} & \textbf{100.0} & \textbf{93.7} & \textbf{82.0} & 55.0 & \textbf{14.3} & 59.2 & \textbf{85.3} & \textbf{73.7} \\
\midrule
\multicolumn{10}{@{}c@{}}{\textbf{Moonlight-16B-A3B}} \\
\textcolor{gray}{BF16} & \textcolor{gray}{100.0} & \textcolor{gray}{100.0} & \textcolor{gray}{99.0} & \textcolor{gray}{97.0} & \textcolor{gray}{100.0} & \textcolor{gray}{98.0} & \textcolor{gray}{99.5} & \textcolor{gray}{76.5} & \textcolor{gray}{96.2} \\
RTN & \textbf{100.0} & 99.0 & 93.0 & 80.0 & 73.0 & 49.0 & 85.8 & 81.2 & 82.6 \\
QuaRot & \textbf{100.0} & \textbf{100.0} & 85.0 & 79.0 & 88.0 & 61.0 & 91.0 & 84.8 & 86.1 \\
KVarN & \textbf{100.0} & 98.0 & 83.0 & 69.0 & 77.0 & 73.0 & 93.8 & 89.8 & 85.4 \\
OSCAR & \textbf{100.0} & \textbf{100.0} & \textbf{96.0} & 92.0 & \textbf{97.0} & 78.0 & 96.2 & \textbf{94.2} & 94.2 \\
\rowcolor{rolaqshade}
\tablemethod{} & \textbf{100.0} & \textbf{100.0} & 94.3 & \textbf{93.3} & 93.0 & \textbf{86.7} & \textbf{98.0} & 93.3 & \textbf{94.8} \\
\bottomrule
\end{tabularx}

\caption{Retrieval accuracy (\%) on the eight RULER retrieval tasks with
4 sink and 16 recent tokens retained in BF16.
Avg. denotes the task mean; bold marks the best 2-bit scores
at the displayed precision.}
\label{tab:ruler}
\end{table}

\subsection{Long-Context Retrieval}
We evaluate eight RULER retrieval tasks at 16K on DeepSeek-V2-Lite
and 8K on Moonlight (Table~\ref{tab:ruler}), their longest evaluated
contexts in this study.
Both exceed the maximum calibration length of 4K.
\method{} leads the 2-bit methods on both models, reaching 73.7\%
on DeepSeek-V2-Lite, 6.3 points above the strongest baseline.
On Moonlight, it reaches 94.8\%, compared with BF16's 96.2\%.
DeepSeek-V2-Lite's MK2 accuracy also improves over OSCAR from
\nDSRulerMultiKeyOSCAR{} to \nDSRulerMultiKeyOurs{}.
These results demonstrate transfer beyond the calibration length;
Section~\ref{sec:why} examines response sensitivity up to 32K.

\section{Ablations and Analysis}
\label{sec:analysis}
\label{sec:ablation}
We examine how rotation, range, and calibration choices affect
accuracy and serving efficiency.

\subsection{Calibration Ablations}
\label{sec:components}
\begin{table}[t]
\centering
\small
\setlength{\tabcolsep}{3pt}
% Generated from final_ablations.json and ablation_score_updates.json.
\begin{tabular*}{\linewidth}{@{\extracolsep{\fill}}lcccc@{}}
\toprule
 & \multicolumn{2}{c}{DeepSeek} & \multicolumn{2}{c}{Moonlight} \\
\cmidrule(lr){2-3}\cmidrule(lr){4-5}
\multicolumn{5}{@{}l}{\textit{(a) Calibration components}} \\
Configuration & MMLU & MBPP & MMLU & MBPP \\
\midrule
Initialization & 52.0 & 39.4 & 67.0 & 48.4 \\
Rotation only & 54.5 & 41.6 & 67.2 & 48.8 \\
Range only & 52.3 & 40.0 & 67.4 & 48.4 \\
Rotation + Range & \textbf{55.9} & \textbf{44.0} & \textbf{67.9} & \textbf{49.2} \\
\midrule
\multicolumn{5}{@{}l}{\textit{(b) Calibration objective}} \\
Objective & \multicolumn{2}{c}{Avg.} & \multicolumn{2}{c}{Avg.} \\
\cmidrule(lr){2-3}\cmidrule(lr){4-5}
KL only & \multicolumn{2}{c}{47.5} & \multicolumn{2}{c}{62.0} \\
Attention-output only & \multicolumn{2}{c}{47.7} & \multicolumn{2}{c}{62.2} \\
KL + attention-output & \multicolumn{2}{c}{\textbf{48.1}} & \multicolumn{2}{c}{\textbf{62.6}} \\
\bottomrule
\end{tabular*}

\caption{Ablations of calibration components and objectives.
(a) MMLU and MBPP accuracy (\%) for different component configurations.
(b) Six-task average accuracy (\%) for different calibration objectives.}
\label{tab:final-components}
\end{table}

\paragraph{Calibration components.}
Joint calibration achieves the highest accuracy on both models
(Table~\ref{tab:final-components}a).
Allowing the range to adapt increases the gains from rotation calibration.
For example, rotation calibration improves DeepSeek-V2-Lite's MBPP accuracy
by 2.2 points with a fixed range and by 4.0 points with a learned range.
These results support adapting the quantizer jointly with the latent
representation.

\paragraph{Calibration objective.}
With both components enabled, the combined objective yields the highest
six-task mean on both models
(Table~\ref{tab:final-components}b).
Attention-output supervision alone slightly outperforms KL alone, while
combining them raises the means to \abDSFullAvg{} on DeepSeek-V2-Lite and
\abMLFullAvg{} on Moonlight.
These gains support supervising both final predictions and per-layer
attention outputs.
We use $\lambda=0.1$ for the combined objective.

\subsection{Rotation Search Space}
\label{sec:geometry-analysis}
\begin{table}[t]
\centering
\small
\setlength{\tabcolsep}{3pt}
% Generated from snapshot.json, final_ablations.json and ablation_score_updates.json.
\begin{tabular*}{\linewidth}{@{\extracolsep{\fill}}lccccc@{}}
\toprule
 & & \multicolumn{2}{c}{DeepSeek} & \multicolumn{2}{c}{Moonlight} \\
\cmidrule(lr){3-4}\cmidrule(lr){5-6}
Rotation & Params. & $S\downarrow$ & Avg.$\uparrow$ & $S\downarrow$ & Avg.$\uparrow$ \\
\midrule
Full (Exp.) & $d^2$ & 0.664 & 47.5 & 0.559 & 62.0 \\
Full (Cayley) & $d^2$ & 0.661 & 47.5 & 0.592 & \textbf{62.3} \\
Random subspaces & $d^2/2$ & 0.628 & 47.4 & 0.550 & 61.9 \\
$H_K{-}H_V$ & $d^2/2$ & 0.624 & 47.1 & 0.579 & 62.2 \\
$H_K{+}H_V$ & $d^2/2$ & \textbf{0.622} & \textbf{47.7} & \textbf{0.508} & \textbf{62.3} \\
\bottomrule
\end{tabular*}

\caption{Rotation search spaces.
Params. counts trainable rotation parameters per layer;
$S$ is validation KL relative to OSCAR.
Avg. is six-task mean accuracy (\%).}
\label{tab:ablation}
\end{table}

\begin{figure}[t]
\centering
\includegraphics[width=\columnwidth]{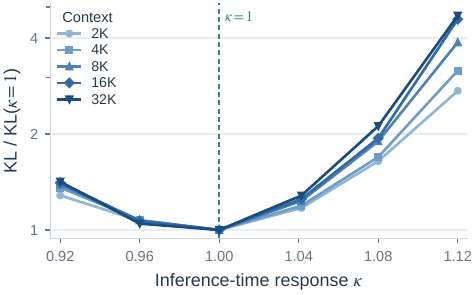}
\caption{Effect of quantizer response on DeepSeek-V2-Lite.
Prediction KL to BF16 is normalized by its value at $\kappa=1$ for each
context length.
Using a checkpoint calibrated with full rotation and unit response,
we vary the gain at inference while keeping the rotation and thresholds
fixed.}
\label{fig:range}
\end{figure}

The $H_K+H_V$ partition uses half the trainable rotation parameters
of full exponential and Cayley rotations while matching or improving
their six-task accuracy (Table~\ref{tab:ablation}).
It also reduces validation KL by 6.3\% and 9.1\% relative to full
exponential rotation on DeepSeek-V2-Lite and Moonlight.
The random control preserves block sizes, initialization, and calibration
budget but replaces the K/V eigenbasis with a random orthogonal basis.
The K/V basis improves both KL and accuracy over this control.
These comparisons support a compact rotation search guided by
MLA's two projection paths (Appendix~\ref{app:geometry}).
The learned blocks export as a dense rotation, allowing the compact
search space to use the same inference kernels as full rotations.

\subsection{Quantizer Response}
\label{sec:why}

Range calibration controls the shared signal used by both attention paths.
Under the Gaussian reference, minimizing reconstruction error can favor
attenuation (Appendix~\ref{app:response}).
Coupling gain to threshold improves MiniCPM3's normalized validation KL
from 0.606 with free gain to 0.527 within the same $H_K+H_V$ search
space (Appendix~\ref{app:range-results}).
The coupled setting also gives more consistent KL across the three
calibration seeds.

We then examine response sensitivity by fixing a DeepSeek-V2-Lite
checkpoint calibrated with $\kappa=1$ and varying its gain at inference.
Figure~\ref{fig:range} shows the lowest prediction KL at $\kappa=1$
across contexts from 2K to 32K.
Amplification becomes more costly as context grows.
Together, these results support adapting quantization boundaries while
controlling the reconstructed signal shared by attention matching
and value readout.

\subsection{Token Protection}
\label{sec:windows}
\begin{figure}[t]
\centering
\includegraphics[width=\columnwidth]{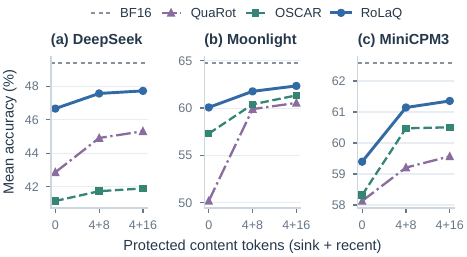}
\caption{Effect of token protection on six-task mean accuracy (\%).
The horizontal axis gives the number of sink + recent content tokens
retained in BF16.}
\label{fig:windows}
\end{figure}

% Declare this wide float early so it appears beside the serving discussion.
\begin{figure*}[t]
\centering
\includegraphics[width=\textwidth]{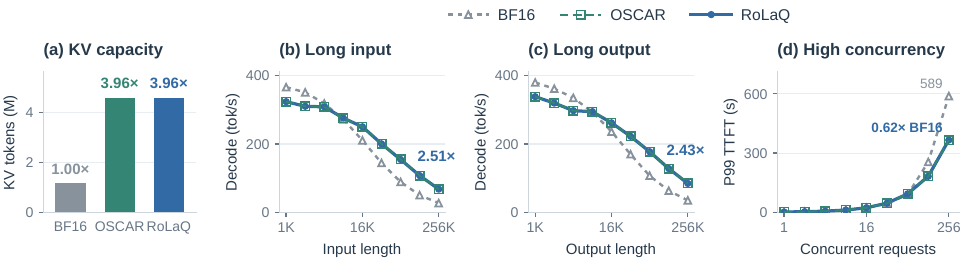}
\caption{Serving performance of DeepSeek-V2-Lite on an A100 80GB GPU.
(a) KV token capacity.
(b,c) Decode throughput for long inputs and successive output intervals,
with four concurrent requests.
(d) P99 time to first token (TTFT) under increasing concurrency.
OSCAR and \method{} share the INT2 kernels, so their overlapping
curves use timings measured with \method{} under the same workloads.}
\label{fig:throughput}
\end{figure*}

The advantage of \method{} is largest without token protection:
it exceeds the strongest baseline by
\nCPMNowinGainBest{}--\nDSNowinGainBest{} points in six-task mean accuracy
across the three models (Table~\ref{tab:main}).
With every content token quantized, these gains highlight the accuracy
preserved by the learned latent quantization itself.
We reuse the same calibrated checkpoints across protection budgets.
Retaining sink and recent tokens in BF16 further improves \method{},
which maintains the highest mean across all three protection budgets
(Figure~\ref{fig:windows}).
Latent calibration improves the low-bit representation, while token
protection retains selected tokens at high precision; their benefits
can therefore be combined.
With only 4 sink and 8 recent tokens protected, \method{} already exceeds
the strongest baseline using 4 sink and 16 recent tokens on each model
(Tables~\ref{tab:s4r8} and~\ref{tab:main}).

\subsection{Serving Efficiency}
\label{sec:efficiency}

We integrate \method{} into SGLang \citep{sglang} with Triton kernels
for packed INT2 cache writes and attention.
We fold read-side rotations into projections and apply thresholds
and gains during cache writes (\S\ref{sec:calibration}).
We benchmark DeepSeek-V2-Lite on one A100 80GB against SGLang's
default BF16 implementation with identical workloads and memory
budgets (Appendix~\ref{app:serving}).

Under the same memory budget, \method{} and OSCAR each accommodate
4.56M KV tokens, compared with 1.15M for BF16, providing $3.96\times$
the capacity (Figure~\ref{fig:throughput}(a)).
Decode throughput reaches $2.51\times$ BF16 with 256K input tokens
and $2.43\times$ during the 128K--256K output interval
(Figure~\ref{fig:throughput}(b,c)).
At 256 concurrent requests, P99 TTFT falls by 37.8\%, from 588.6\,s
to 366.0\,s (Figure~\ref{fig:throughput}(d)).
These measurements demonstrate greater KV capacity, faster decoding,
and lower first-token tail latency in a practical serving system.

\section{Related Work}

\paragraph{KV cache quantization.}
KV cache quantization reduces the memory cost of long-context inference
by storing cached states at lower precision.
KIVI adapts quantization granularity to key and value statistics, while
KVQuant combines non-uniform quantization with outlier handling
\citep{kivi,kvquant}.
GEAR combines low-bit quantization with low-rank and sparse components to
compensate errors \citep{gear}, while ZipCache assigns higher precision
to salient tokens \citep{zipcache}.
Retaining recent tokens or attention sinks in full precision provides
another way to control error \citep{kivi,rotatekv}.
SnapMLA combines FP8 MLA caches with hardware-aware decoding
while retaining RoPE keys in high precision \citep{snapmla}.

\paragraph{Rotation and calibration.}
Beyond the quantizer itself, representation transforms help control outliers
at low bitwidths while preserving full-precision computation.
QuaRot uses Hadamard rotations, while SpinQuant learns rotations through
calibration \citep{quarot,spinquant}.
RotateKV and KVarN adapt KV caches through outlier-aware rotations and
variance normalization, respectively \citep{rotatekv,kvarn}.
OmniQuant learns clipping and equivalent transformations with pretrained
weights frozen, motivating joint calibration of representations and
quantization ranges \citep{omniquant}.

Recent methods further tailor calibration to how cache errors affect attention.
OSCAR derives rotations from attention-aware covariance statistics and
supports MLA \citep{oscar}, while OptR learns per-head key and value
rotations through attention-output reconstruction after the output
projection \citep{optr}.
QuantMLA learns offline-fusible transformations for content and RoPE caches,
using attention-output and positional-QK reconstruction objectives,
respectively \citep{quantmla}.
\method{} uses MLA's K/V projection structure to guide joint calibration
of the shared latent rotation and quantization range.

\section{Conclusion}
We presented \method{} for joint calibration of MLA's latent rotation
and quantization range at two bits.
K/V projection geometry provides a compact rotation search space,
while coupled thresholds and gains control reconstruction response.
Experiments show higher average accuracy across three MLA models and
better long-context retrieval, with further gains from token protection.
Our SGLang implementation increases KV capacity and throughput
while reducing latency at high concurrency.
\section*{Limitations}
Further compression of MLA caches must account for both content and
position.
\method{} quantizes the shared content latent while retaining the RoPE
key in BF16, leaving positional keys and group metadata as a substantial
share of the remaining cache.
Extending these savings calls for coordinated content and positional
precision while preserving attention's ability to distinguish positions.
The serving benefits of low-bit caches also depend on hardware and
workload.
Bandwidth savings are most valuable when reading a long history dominates
decoding.
At shorter contexts, quantization and unpacking overheads can offset them.
Our serving measurements use DeepSeek-V2-Lite on an A100.
Extending these benefits calls for kernels tailored to the target
hardware and workload.

\bibliography{references}
\clearpage
\appendix
\section{Implementation and Evaluation Details}
\label{app:protocol}
\subsection{Models and Evaluation}
The three models let us examine shared-latent quantization across both
mixture-of-experts and dense architectures (Table~\ref{tab:model-config}).
DeepSeek-V2-Lite and Moonlight each have 16B total parameters;
Moonlight follows the DeepSeek-V3 architecture.
MiniCPM3 is a dense 4B model with a smaller latent and more layers.
We use lm-evaluation-harness \citep{lmeval} with the few-shot settings
in Table~\ref{tab:eval-config} and greedy decoding for generation tasks.

\begin{table}[!htbp]
\centering
\footnotesize
\begin{tabular*}{\linewidth}{@{\extracolsep{\fill}}llrrr@{}}
\toprule
Model & Type & Layers & $d$ & $d_r$ \\
\midrule
DeepSeek-V2-Lite & MoE & 27 & 512 & 64 \\
Moonlight-16B-A3B & MoE & 27 & 512 & 64 \\
MiniCPM3-4B & Dense & 62 & 256 & 32 \\
\bottomrule
\end{tabular*}
\caption{Model configurations; $d$ and $d_r$ are the content-latent and RoPE-key
dimensions.}
\label{tab:model-config}
\end{table}

\begin{table}[!htbp]
\centering
\small
\begin{tabular*}{\linewidth}{@{\extracolsep{\fill}}lrl@{}}
\toprule
Benchmark & Shots & Metric \\
\midrule
MMLU & 5 & Accuracy \\
CMMLU & 5 & Accuracy \\
TriviaQA & 5 & Exact match \\
GSM8K & 5 & Exact match \\
HumanEval & 0 & pass@1 \\
MBPP & 3 & pass@1 \\
\bottomrule
\end{tabular*}
\caption{Downstream evaluation settings.}
\label{tab:eval-config}
\end{table}

The benchmarks cover both answer selection and free-form generation.
MMLU and CMMLU report accuracy weighted by subject size;
TriviaQA uses exact match, and GSM8K uses flexible answer extraction.
HumanEval and MBPP report greedy pass@1.
MiniCPM3 uses its chat template, with HumanEval code extracted from
the generated response.
For MBPP, we truncate code at the first new example or trailing
assertion block and remove test calls before rescoring all methods
with a common evaluator.

Our long-context evaluation focuses on RULER's eight retrieval tasks:
three single-key, three multi-key, one multi-value, and one multi-query
task.
This set tests whether the cache preserves the information needed to
recover a specified item as the surrounding context grows.
Each task contains 100 examples scored with the official evaluation
functions.

\subsection{Baseline Implementations}
\label{app:baselines}
We compare the methods at the same point in the MLA computation:
the shared content latent before the K/V up-projections.
The RoPE key remains in BF16 for every method.
The group size is 128; RTN, QuaRot, and OSCAR group latent coordinates within
each token, whereas KVarN groups tokens for channel-wise quantization.
The formulas below describe the transformations used in our MLA evaluation.
We write $\widehat Q=\operatorname{DQ}\circ Q$ for quantization
followed by reconstruction.

\paragraph{RTN.}
For a group $x$, let $m_x=\min_i x_i$ and $M_x=\max_i x_i$.
With step size $\Delta_x=(M_x-m_x)/3$, the affine min--max quantizer is
\begin{equation}
\begin{aligned}
q_x&=\operatorname{clip}_{[0,3]}\!\left(
\operatorname{round}\frac{x-m_x}{\Delta_x}\right),\\
\widehat Q_{\mathrm{mm}}(x)&=m_x+\Delta_xq_x.
\end{aligned}
\label{eq:rtn-baseline}
\end{equation}
Constant groups are reconstructed exactly.
RTN applies $\widehat Q_{\mathrm{mm}}$ directly to each content-latent group.

\paragraph{QuaRot.}
Following the cache rotation in QuaRot \citep{quarot}, we use a block-diagonal
matrix $H_{128}$ of normalized 128-dimensional Hadamard transforms:
\begin{equation}
\hat c_t=\widehat Q_{\mathrm{mm}}(c_tH_{128})H_{128}^{\top}.
\end{equation}
The Hadamard transform mixes coordinates before rounding, providing
a data-independent rotation reference with the backbone frozen.

\paragraph{KVarN.}
KVarN \citep{kvarn} combines a normalized Hadamard rotation $H$ with variance
balancing across tokens and latent channels.
For a tile $X$ of up to 128 tokens, let $Z=(XH)^{\top}$.
Diagonal scales $D_c$ and $D_t$ balance its channel and token axes, respectively:
\begin{equation}
\begin{aligned}
\widetilde Z&=D_c^{-1}ZD_t^{-1},\\
\hat X&=\bigl(D_c\widehat Q_{\mathrm{mm}}(\widetilde Z)D_t\bigr)^{\top}H^{\top}.
\end{aligned}
\end{equation}
We use 16 alternating variance-normalization iterations and apply
$\widehat Q_{\mathrm{mm}}$ to each row of $\widetilde Z$ across tokens.
Padding is excluded from the statistics.
We also quantize partial tiles; single-token updates use Hadamard
rotation followed by RTN over groups of latent coordinates.

\paragraph{OSCAR.}
We use the official MLA recipe of OSCAR \citep{oscar}, with the per-layer
rotation $R_0$ fitted on Pile text by its calibration code.
Let $\widehat Q_{T_0,g_0}$ denote its group-normalized four-level reconstruction in
Eq.~\ref{eq:quantizer}, with fixed threshold $T_0=0.9810653$ and gain $g_0=1.16$:
\begin{equation}
\hat c_t=\widehat Q_{T_0,g_0}(c_tR_0)R_0^{\top}.
\end{equation}
We also use $R_0$ to initialize \method{}.

\subsection{Calibration and Storage}
\label{app:calibration}
Calibration adapts the cache representation while leaving the language
model unchanged.
All three models use the settings in Table~\ref{tab:calibration-config}.
The Pile training text is disjoint from the text used to fit OSCAR's
initial rotation.
We update only the rotation and thresholds, with the teacher and
backbone frozen.
No tokens are protected during calibration, so the learned parameters
are exposed to quantization throughout the sequence.
We reuse these checkpoints across all protection budgets.

\begin{table}[!htbp]
\centering
\footnotesize
\begin{tabular*}{\linewidth}{@{\extracolsep{\fill}}ll@{}}
\toprule
Setting & Value \\
\midrule
Calibration data & Pile, 1,430,800 tokens \\
Sequence length & 512--4,096 tokens \\
Optimizer / updates & Adam / 427 \\
Rotation learning rate & $10^{-3}$ \\
Threshold learning rate & $5\times10^{-3}$ \\
Learning-rate schedule & Cosine \\
Gradient-norm clipping & 1.0 \\
Auxiliary weight $\lambda$ & 0.1 \\
Response $\kappa$ & 1 \\
Quantization group size & 128 \\
Protected training tokens & 0 \\
Random seeds & 42, 43, 44 \\
\bottomrule
\end{tabular*}
\caption{Calibration hyperparameters.}
\label{tab:calibration-config}
\end{table}

\paragraph{Selecting a checkpoint.}
We select checkpoints by prediction fidelity on held-out Pile text.
For each context setting, $r$ denotes the teacher-forced KL to BF16
divided by OSCAR's KL on the same text.
The selection score is a weighted average of the KL ratios for
short, 2K, and 4K contexts:
\begin{equation}
S=0.5\,r_{\mathrm{short}}+0.25\,r_{\mathrm{2K}}
  +0.25\,r_{\mathrm{4K}}.
\label{eq:selection-score}
\end{equation}
We keep the checkpoint with the lowest $S$.
Thus, $S<1$ indicates lower prediction divergence than OSCAR,
and selection reflects behavior across several context lengths.

\paragraph{Quantization grid.}
We use groups of $G=128$ coordinates and OSCAR's four-level affine grid,
with $a_{\mathrm{out}}=1.5095586$ and
$a_{\mathrm{in}}=a_{\mathrm{out}}/3$.
The outer level and initial threshold $T_0=0.9810653$ follow its
Gaussian Lloyd--Max construction; the inner levels give an equally
spaced reconstruction grid.
Each layer learns its threshold and derives the gain from
$g=1/C(T)$, starting at $g\approx1.114$ for the initial threshold.

\paragraph{Storage accounting.}
The cache stores three kinds of information: packed content codes,
group statistics, and the RoPE key.
With two bits per content coordinate, two FP32 scalars per group,
and a BF16 RoPE key, an unprotected token requires
$d/4+8d/G+2d_r$ bytes per layer.
This gives 288 bytes for DeepSeek-V2-Lite and Moonlight,
and 144 bytes for MiniCPM3.
Protected tokens additionally require BF16 content storage.
The allocation measurements in Appendix~\ref{app:serving} include the
packed buffers, group metadata, and token-protection buffers.

\FloatBarrier
\section{Supplementary Method Analysis}
The method combines a choice of rotation space with control over
quantizer response.
We first describe the observations behind these choices, then make the
rotation variants explicit and derive the response constraint.

\subsection{Motivating Observations}
\label{app:observations}
\begin{figure*}[!t]
\centering
\includegraphics[width=\textwidth]{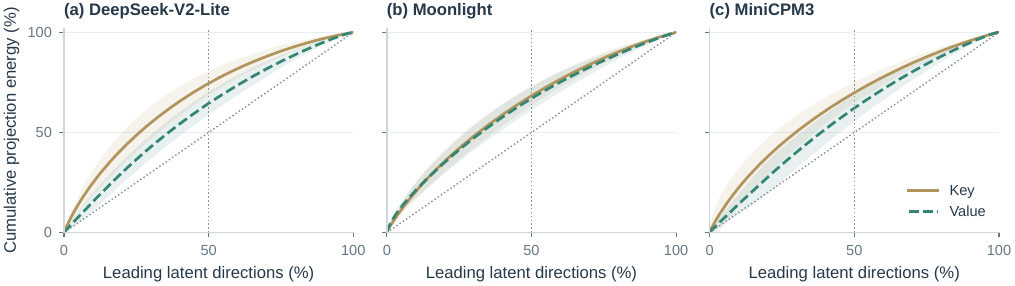}
\caption{Cumulative K/V projection energy in the common joint eigenbasis.
Solid and dashed curves denote keys and values, respectively;
shading shows one standard deviation across layers.
The diagonal gives the isotropic reference.
The vertical line marks the equal-dimensional split used in
Figure~\ref{fig:motivation}a.}
\label{fig:motivation-spectra}
\end{figure*}

The two observations separate direction from scale.
Projection energy measures how strongly latent directions affect
the K/V projections, while the gain intervention changes reconstruction strength
without changing the rotation or the assignments.
\paragraph{Projection energy.}
For each layer, we extract the native, head-interleaved key and value blocks of
the frozen up-projection and form $H_K,H_V$ as in Eq.~\ref{eq:kv-geometry}.
Let $u_1,\ldots,u_d$ be the eigenvectors of $H_K+H_V$, ordered by descending
eigenvalue.
For $X\in\{K,V\}$, the fraction of projection energy in the first $r$ directions is
\begin{equation}
F_X(r)=\frac{\sum_{i=1}^{r}\|W_Xu_i\|_2^2}{\|W_X\|_F^2}.
\label{eq:projection-energy}
\end{equation}
Figure~\ref{fig:motivation}a reports $100F_X(d/2)$, averaged equally over all
27 DeepSeek-V2-Lite layers, 27 Moonlight layers, and 62 MiniCPM3 layers.
The error bars are sample standard deviations across layers.
The same subspace is used for the key and value measurements within each layer.
All matrix products and eigendecompositions use FP64 on the original stored
weights.
Figure~\ref{fig:motivation-spectra} shows the full cumulative energy
profiles for both projections.

\paragraph{Gain intervention.}
Figure~\ref{fig:motivation}b uses the OSCAR rotation $R_0$ before learned
refinement and its threshold $T_0=0.981065$ in every layer.
We vary the gain uniformly across layers, holding both $R_0$ and $T_0$ fixed.
The six evaluated gains are approximately
$\{1.025,1.069,1.114,1.160,1.203,1.248\}$; the original gain is $g_0=1.16$.
All content tokens are quantized; RoPE keys remain in BF16.
We use 12 natural documents, two each from PG-19, Wikipedia, EuroParl,
HackerNews, arXiv, and Pile-CC, with 8,192 tokens of context.
Each document contributes the mean
$D_{\mathrm{KL}}(p_{\mathrm{BF16}}\|p_{\mathrm{INT2}})$ over its final 1,024
next-token predictions.
Each plotted value is the mean over documents divided by that model's mean at
$g_0$; the curves connect the six measured settings.

\subsection{Rotation Search Spaces}
\label{app:geometry}
All variants start from the same covariance-aware rotation $R_0$.
They differ in which directions can mix during refinement and in how
the orthogonal update is parameterized.
Writing these choices explicitly separates the effect of the K/V
geometry from the size of the trainable search space.

\paragraph{Full exponential rotation.}
A dense generator $A\in\R^{d\times d}$ defines
\begin{equation}
R_{\mathrm{full}}=R_0e^\Omega,\qquad \Omega=A-A^\top.
\end{equation}
Every latent direction can mix with every other direction.
This provides the unrestricted exponential reference for the
structured variants.

\paragraph{Cayley rotation.}
Using the same dense skew-symmetric generator, Cayley replaces the
matrix exponential with
\begin{equation}
R_{\mathrm{Cayley}}=R_0(I-\Omega)(I+\Omega)^{-1}.
\end{equation}
The trainable parameter count remains $d^2$, so this comparison changes
the orthogonal map while keeping the generator unrestricted.

\paragraph{Joint-energy blocks: $H_K+H_V$.}
The structured variants use
$B=\operatorname{diag}(e^{S_1},e^{S_2})$, where
$S_i=A_i-A_i^\top$ and $A_i\in\R^{d/2\times d/2}$.
For the joint-energy basis,
\begin{equation}
\begin{aligned}
R_0^\top(H_K+H_V)R_0&=U_+\Lambda_+U_+^\top,\\
R_+&=R_0U_+BU_+^\top.
\end{aligned}
\end{equation}
We order eigenvalues from largest to smallest and split $U_+$ into
two equal halves.
These blocks separate high and low joint projection energy,
while each still contributes to both keys and values.

\paragraph{K/V-contrast blocks: $H_K-H_V$.}
Replacing the sum by the difference gives
\begin{equation}
\begin{aligned}
R_0^\top(H_K-H_V)R_0&=U_-\Lambda_-U_-^\top,\\
R_-&=R_0U_-BU_-^\top.
\end{aligned}
\end{equation}
Here the signed eigenvalues measure the relative strength of the
key and value projections along each direction.
The leading half is more key-dominant; the trailing half is
more value-dominant.

\paragraph{Random subspaces.}
The random control uses the same block sizes but replaces the
K/V basis with a fixed random orthogonal basis:
\begin{equation}
R_{\mathrm{rand}}=R_0U_{\mathrm{rand}}BU_{\mathrm{rand}}^\top.
\end{equation}
It therefore retains the $d^2/2$ trainable parameters of the spectral
variants while removing their K/V-based grouping.
All generators are initialized at zero, recovering $R_0$,
and all variants export a dense $d\times d$ rotation for inference.
The random control uses seed 42; the other variants use three seeds.

\subsection{Response of the Four-Level Quantizer}
\label{app:response}
Decision thresholds and reconstruction levels determine a scalar
quantizer's distortion \citep{lloyd1982,max1960}.
To isolate the scale change introduced by quantization, we separate
the part of the reconstruction aligned with the input from its residual.
This gives the linear response used in \S\ref{sec:range}.
For $x\sim\mathcal{N}(0,1)$, let $\phi$ denote the standard normal density.
Symmetry of the code in Eq.~\ref{eq:quantizer} gives
\begin{equation}
\begin{aligned}
\E[x\,a_T(x)]={}&2a_{\mathrm{in}}\!\int_0^T\! x\phi(x)\,dx\\
&+2a_{\mathrm{out}}\!\int_T^\infty\! x\phi(x)\,dx\\
={}&2a_{\mathrm{in}}\big[\phi(0)-\phi(T)\big]+2a_{\mathrm{out}}\,\phi(T),
\end{aligned}
\end{equation}
which is $C(T)$, using $\int x\phi(x)\,dx=-\phi(x)$.
Since $\E[x^2]=1$, the least-squares coefficient of $g\,a_T(x)$ on $x$ is
$\kappa=g\,C(T)$, and the residual $\epsilon=g\,a_T(x)-\kappa x$ satisfies
$\E[x\epsilon]=0$ \citep{bussgang1952,bussgang}.
Setting $\kappa=1$ preserves the component aligned with the input,
while $\epsilon$ captures the remaining quantization noise.

For a fixed threshold, the gain that minimizes the Gaussian reconstruction error is
$g_{\mathrm{MSE}}=C(T)/\E[a_T(x)^2]$, which gives
$\kappa_{\mathrm{MSE}}=C(T)^2/\E[a_T(x)^2]\approx0.88$ near $T=1$.
Thus, reducing reconstruction error can favor a weaker reconstructed
signal.
Coupling $g$ to $T$ through $1/C(T)$ instead anchors the aligned
component while allowing the decision boundary to adapt.
Section~\ref{sec:why} tests how this response affects prediction fidelity
across context lengths.

We bound $T\in[0.5,1.25]$ so that $g=1/C(T)\lesssim1.3$.
The official setting $(T,g)=(0.981,1.16)$ gives $\kappa=\nKappaOfficial$, and $\kappa=1$
at the official threshold corresponds to $g=1.114$.

\FloatBarrier
\section{Additional Results}
These results expand the main comparisons along three dimensions:
token-protection budget, calibration seed, and range calibration.
We then examine the learned thresholds and concrete model responses
to connect the aggregate scores with the behavior of the cache.

\subsection{Token Protection and Calibration Seeds}
\label{app:windows}
\label{app:seeds}
The intermediate budget of 4 sink and 8 recent tokens fills the gap
between the two settings in Table~\ref{tab:main}.
Table~\ref{tab:s4r8} reports all six tasks at this budget,
where \method{} again has the highest mean among the compared
2-bit methods on all three models.
Together with the main table, it shows how latent calibration and
token protection combine across protection budgets.

\begin{table*}[!t]
\centering
\small
\setlength{\tabcolsep}{5pt}
\begin{tabular}{lccccccc}
\toprule
\multirow{2}{*}{\textbf{Method}} & \multicolumn{3}{c}{\textbf{Knowledge \& QA}} & \multicolumn{1}{c}{\textbf{Math}} & \multicolumn{2}{c}{\textbf{Code}} & \multirow{2}{*}{\textbf{Avg.}} \\
\cmidrule(lr){2-4}\cmidrule(lr){5-5}\cmidrule(lr){6-7}
 & MMLU & CMMLU & TriviaQA & GSM8K & HumanEval & MBPP & \\
\midrule
\multicolumn{8}{c}{\textbf{DeepSeek-V2-Lite}} \\
\textcolor{gray}{BF16} & \textcolor{gray}{58.0} & \textcolor{gray}{64.1} & \textcolor{gray}{65.2} & \textcolor{gray}{37.4} & \textcolor{gray}{28.0} & \textcolor{gray}{43.4} & \textcolor{gray}{49.4} \\
RTN & 43.0 & 49.6 & 55.0 & 13.3 & 18.9 & 26.0 & 34.3 \\
QuaRot & 53.1 & 60.1 & 62.4 & 30.3 & \textbf{27.4} & 36.2 & 44.9 \\
KVarN & 52.8 & 59.0 & 62.6 & 29.5 & 24.4 & 39.8 & 44.7 \\
OSCAR & 49.9 & 57.4 & 61.2 & 18.4 & 25.6 & 37.8 & 41.7 \\
\tablemethod{} & \textbf{55.5} & \textbf{62.1} & \textbf{64.4} & \textbf{33.8} & \textbf{27.4} & \textbf{42.1} & \textbf{47.6}\std{0.6} \\
\midrule
\multicolumn{8}{c}{\textbf{Moonlight-16B-A3B}} \\
\textcolor{gray}{BF16} & \textcolor{gray}{70.0} & \textcolor{gray}{78.2} & \textcolor{gray}{66.4} & \textcolor{gray}{73.5} & \textcolor{gray}{47.0} & \textcolor{gray}{53.4} & \textcolor{gray}{64.7} \\
RTN & 65.1 & 71.8 & 62.3 & 60.9 & 39.0 & 47.2 & 57.7 \\
QuaRot & 65.4 & 72.6 & 62.3 & 64.8 & 42.7 & \textbf{51.6} & 59.9 \\
KVarN & 63.0 & 67.4 & 57.3 & 61.0 & 41.5 & 43.0 & 55.5 \\
OSCAR & \textbf{67.5} & 74.9 & 63.0 & 68.8 & 40.2 & 48.0 & 60.4 \\
\tablemethod{} & \textbf{67.5} & \textbf{75.7} & \textbf{64.2} & \textbf{70.0} & \textbf{43.9} & 49.3 & \textbf{61.8}\std{0.4} \\
\midrule
\multicolumn{8}{c}{\textbf{MiniCPM3-4B}} \\
\textcolor{gray}{BF16} & \textcolor{gray}{64.4} & \textcolor{gray}{72.9} & \textcolor{gray}{36.2} & \textcolor{gray}{75.7} & \textcolor{gray}{74.4} & \textcolor{gray}{51.8} & \textcolor{gray}{62.6} \\
RTN & 58.9 & 66.9 & 32.1 & 64.1 & 68.9 & 46.8 & 56.3 \\
QuaRot & 62.4 & 70.0 & 34.0 & 70.1 & 70.7 & 48.2 & 59.2 \\
KVarN & 62.8 & 70.8 & 34.4 & 72.3 & 67.7 & 50.4 & 59.7 \\
OSCAR & 63.0 & \textbf{71.4} & 34.5 & \textbf{73.5} & 70.1 & 50.2 & 60.5 \\
\tablemethod{} & \textbf{63.4} & 71.1 & \textbf{34.9} & 73.4 & \textbf{73.4} & \textbf{50.7} & \textbf{61.1}\std{0.1} \\
\bottomrule
\end{tabular}

\caption{Accuracy (\%) with 4 sink and 8 recent BF16 tokens.}
\label{tab:s4r8}
\end{table*}

Tables~\ref{tab:seeds-nowin}, \ref{tab:seeds-s4r8}, and \ref{tab:seeds}
expand the method averages into seeds 42, 43, and 44 for the three
protection settings.
Each seed corresponds to one calibrated checkpoint reused across
protection budgets, so the tables show both calibration variation
and the effect of protecting more tokens.
Standard deviations are computed across seeds for each task;
for Avg., they are computed from the three six-task means.
DS, ML, and CPM denote DeepSeek-V2-Lite, Moonlight, and MiniCPM3.

\begin{table*}[!t]
\centering
\small
\begin{tabular*}{\textwidth}{@{\extracolsep{\fill}}lrccccccc@{}}
\toprule
Model & Seed & MMLU & CMMLU & TriviaQA & GSM8K & HumanEval & MBPP & Avg. \\
\midrule
DS & 42 & 55.1 & 61.5 & 63.8 & 31.6 & 26.8 & 42.4 & 46.9 \\
DS & 43 & 55.3 & 61.5 & 64.1 & 32.5 & 26.2 & 42.2 & 47.0 \\
DS & 44 & 54.8 & 61.6 & 64.3 & 32.3 & 23.2 & 40.8 & 46.2 \\
 & std & 0.21 & 0.04 & 0.24 & 0.47 & 1.96 & 0.87 & 0.44 \\
\midrule
ML & 42 & 67.1 & 74.8 & 63.0 & 65.7 & 38.4 & 47.2 & 59.4 \\
ML & 43 & 67.6 & 74.8 & 63.0 & 67.0 & 42.7 & 48.8 & 60.7 \\
ML & 44 & 67.5 & 75.2 & 63.5 & 66.9 & 40.2 & 47.8 & 60.2 \\
 & std & 0.26 & 0.24 & 0.27 & 0.77 & 2.14 & 0.81 & 0.66 \\
\midrule
CPM & 42 & 63.3 & 71.0 & 35.1 & 70.7 & 68.3 & 50.0 & 59.7 \\
CPM & 43 & 63.0 & 70.9 & 34.3 & 70.4 & 64.0 & 50.0 & 58.8 \\
CPM & 44 & 62.3 & 71.1 & 34.5 & 70.2 & 68.9 & 51.0 & 59.7 \\
 & std & 0.54 & 0.09 & 0.42 & 0.27 & 2.66 & 0.58 & 0.53 \\
\bottomrule
\end{tabular*}

\caption{Per-seed accuracy (\%) without protection.
Avg. averages six tasks; std denotes sample standard deviation.}
\label{tab:seeds-nowin}
\end{table*}

\begin{table*}[!t]
\centering
\small
\begin{tabular*}{\textwidth}{@{\extracolsep{\fill}}lrccccccc@{}}
\toprule
Model & Seed & MMLU & CMMLU & TriviaQA & GSM8K & HumanEval & MBPP & Avg. \\
\midrule
DS & 42 & 55.6 & 62.5 & 64.4 & 34.1 & 29.9 & 43.2 & 48.3 \\
DS & 43 & 55.4 & 61.8 & 64.2 & 33.8 & 26.8 & 41.4 & 47.2 \\
DS & 44 & 55.7 & 62.1 & 64.5 & 33.5 & 25.6 & 41.8 & 47.2 \\
 & std & 0.15 & 0.34 & 0.16 & 0.30 & 2.20 & 0.95 & 0.61 \\
\midrule
ML & 42 & 67.3 & 75.9 & 64.7 & 71.3 & 47.0 & 47.4 & 62.3 \\
ML & 43 & 67.7 & 75.6 & 64.0 & 68.8 & 42.7 & 51.0 & 61.6 \\
ML & 44 & 67.5 & 75.5 & 64.0 & 70.0 & 42.1 & 49.6 & 61.4 \\
 & std & 0.17 & 0.19 & 0.42 & 1.29 & 2.66 & 1.81 & 0.44 \\
\midrule
CPM & 42 & 63.4 & 71.1 & 35.0 & 73.1 & 73.2 & 50.2 & 61.0 \\
CPM & 43 & 63.4 & 71.2 & 34.8 & 73.3 & 73.8 & 51.2 & 61.3 \\
CPM & 44 & 63.5 & 71.1 & 35.0 & 73.8 & 73.2 & 50.6 & 61.2 \\
 & std & 0.06 & 0.04 & 0.12 & 0.39 & 0.35 & 0.50 & 0.15 \\
\bottomrule
\end{tabular*}

\caption{Per-seed accuracy (\%) with 4 sink and 8 recent BF16 tokens.}
\label{tab:seeds-s4r8}
\end{table*}

\begin{table*}[!t]
\centering
\small
\begin{tabular*}{\textwidth}{@{\extracolsep{\fill}}lrccccccc@{}}
\toprule
Model & Seed & MMLU & CMMLU & TriviaQA & GSM8K & HumanEval & MBPP & Avg. \\
\midrule
DS & 42 & 55.9 & 62.5 & 64.8 & 34.6 & 26.8 & 44.0 & 48.1 \\
DS & 43 & 55.5 & 61.6 & 64.8 & 34.3 & 29.3 & 42.8 & 48.0 \\
DS & 44 & 55.9 & 62.1 & 65.0 & 33.6 & 25.0 & 40.6 & 47.0 \\
 & std & 0.25 & 0.45 & 0.14 & 0.54 & 2.14 & 1.72 & 0.60 \\
\midrule
ML & 42 & 67.9 & 76.2 & 65.7 & 70.4 & 46.3 & 49.2 & 62.6 \\
ML & 43 & 67.8 & 76.1 & 65.2 & 70.1 & 44.5 & 51.6 & 62.5 \\
ML & 44 & 68.0 & 75.9 & 64.9 & 71.0 & 42.1 & 49.4 & 61.9 \\
 & std & 0.09 & 0.18 & 0.41 & 0.46 & 2.14 & 1.33 & 0.42 \\
\midrule
CPM & 42 & 63.7 & 71.1 & 35.3 & 74.1 & 76.2 & 50.4 & 61.8 \\
CPM & 43 & 63.4 & 71.4 & 35.2 & 75.1 & 73.2 & 48.8 & 61.2 \\
CPM & 44 & 63.4 & 71.3 & 35.4 & 74.5 & 72.0 & 49.8 & 61.1 \\
 & std & 0.17 & 0.13 & 0.10 & 0.50 & 2.20 & 0.81 & 0.41 \\
\bottomrule
\end{tabular*}

\caption{Per-seed accuracy (\%) with 4 sink and 16 recent BF16 tokens.}
\label{tab:seeds}
\end{table*}

\subsection{Range Calibration on MiniCPM3}
\label{app:range-results}
With the $H_K+H_V$ search space fixed, Table~\ref{tab:selection} compares
freely learned gain with gain coupled to the threshold on MiniCPM3.
Coupling gain to threshold reduces the mean validation KL ratio
from 0.606 to 0.527 and its variation across calibration seeds.
Thus, response control improves prediction fidelity within the
same structured rotation space.

\begin{table*}[!t]
\centering
\begin{minipage}{\columnwidth}
\small
\begin{tabular*}{\linewidth}{@{\extracolsep{\fill}}lc@{}}
\toprule
Range & Validation KL $S\downarrow$ \\
\midrule
Free gain & 0.606\std{0.031} \\
$\kappa=1$ & \textbf{0.527}\std{0.008} \\
\bottomrule
\end{tabular*}

\caption{Range calibration with $H_K+H_V$ on MiniCPM3.
$S$ is validation KL relative to OSCAR (lower is better);
values are mean $\pm$ sample standard deviation over three seeds.}
\label{tab:selection}
\end{minipage}
\end{table*}

\subsection{Learned Thresholds}
\label{app:thresholds}
\begin{figure*}[!t]
\centering
\includegraphics[width=\textwidth]{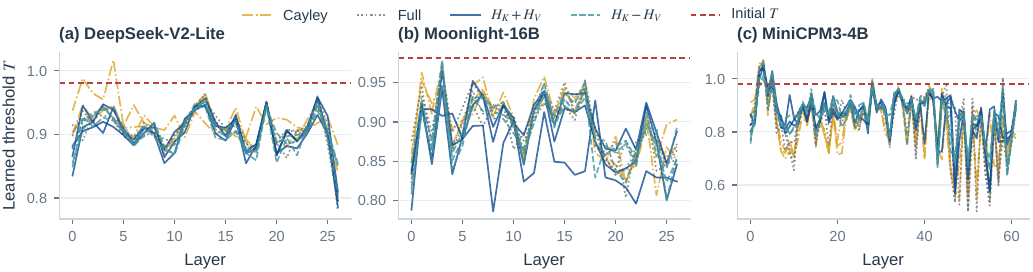}
\caption{Learned per-layer thresholds with $\kappa=1$ for all rotations and seeds.
The dashed line is the Lloyd--Max threshold used by OSCAR.}
\label{fig:thresholds}
\end{figure*}

The response constraint leaves each layer free to choose its
quantization boundary.
Figure~\ref{fig:thresholds} shows the resulting profiles across
rotations and seeds.
Most layers move below the initial Lloyd--Max threshold, with
thresholds concentrated around $T\approx\nDSTKone$,
\nMLTKone{}, and \nCPMTKone{} for the three models.
Lower thresholds narrow the central assignment region, while the
coupled gain preserves unit response.
The layerwise profiles therefore show how the quantizer adapts its
assignments under a shared response constraint.

\subsection{Case Study}
\label{app:cases}
Tables~\ref{tab:case-gsm} and \ref{tab:case-ruler} present four
selected examples from DeepSeek-V2-Lite.
With the same token protection budget, \method{} preserves the
correct BF16 answers while both QuaRot and OSCAR fail.

In Table~\ref{tab:case-gsm}(a), both baselines divide the chips
in one bag among five people, overlooking the second bag.
\method{} counts both bags and obtains 22 chips per person.
In (b), both baselines return the total requirement of 600 feet;
\method{} subtracts the 6 feet already available and answers 594.
Across all \nCaseTotal{} GSM8K problems, \method{} solves
\nCaseWins{} that OSCAR misses and misses \nCaseLosses{} that OSCAR
solves.

In 16K retrieval (Table~\ref{tab:case-ruler}), QuaRot changes two
digits of the value requested by a single-key query,
while OSCAR returns a different number.
For four queried keys, OSCAR repeats two values and omits the others;
QuaRot repeats an unrelated number.
\method{} retrieves the correct value or complete set in each case.

% Keep all four qualitative examples on the same page.
\begin{table*}[!t]
\centering
\small
\begin{tabular}{@{}p{0.12\textwidth}p{0.83\textwidth}@{}}
\toprule
\multicolumn{2}{@{}l@{}}{\textbf{(a) Counting both bags}} \\
\midrule
Question & Mitchell is making nachos for his family. He buys two bags of chips with 55 chips each. If his family has five members, how many chips does each person get if they all get the same number? \\
Answer & 22 \\
\midrule
BF16 & 22 \\
\midrule
QuaRot & \textcolor{red!60!black}{Each person gets 55 / 5 = \textless{}\textless{}55/5=11\textgreater{}\textgreater{}11 chips.\newline \#\#\#\# 11} \\
\midrule
OSCAR & \textcolor{red!60!black}{There are 55 chips in each bag.\newline If there are five people, each person gets 55 / 5 = \textless{}\textless{}55/5=11\textgreater{}\textgreater{}11 chips.\newline \#\#\#\# 11} \\
\midrule
\tablemethod{} & \textcolor{blue!60!black}{Mitchell buys 2 * 55 = \textless{}\textless{}2*55=110\textgreater{}\textgreater{}110 chips.\newline There are 5 members in his family, so each person gets 110 / 5 = \textless{}\textless{}110/5=22\textgreater{}\textgreater{}22 chips.\newline \#\#\#\# 22} \\
\midrule
\multicolumn{2}{@{}l@{}}{\textbf{(b) Subtracting the available line}} \\
\midrule
Question & The caretaker of the docks needs to buy some new line. He wants 3 feet of line for every foot of dock. Right now, there is 200 feet of dock, and he has 6 feet of new line. How many feet of line does he need to buy in total? \\
Answer & 594 \\
\midrule
BF16 & 594 \\
\midrule
QuaRot & \textcolor{red!60!black}{He needs 3*200 = \textless{}\textless{}3*200=600\textgreater{}\textgreater{}600 feet of line.\newline \#\#\#\# 600} \\
\midrule
OSCAR & \textcolor{red!60!black}{He needs 200*3 = \textless{}\textless{}200*3=600\textgreater{}\textgreater{}600 feet of line.\newline \#\#\#\# 600} \\
\midrule
\tablemethod{} & \textcolor{blue!60!black}{He needs 3 feet of line for every foot of dock, so he needs 3*200 = \textless{}\textless{}3*200=600\textgreater{}\textgreater{}600 feet of line.\newline He has 6 feet of line, so he needs to buy 600-6 = \textless{}\textless{}600-6=594\textgreater{}\textgreater{}594 feet of line.\newline \#\#\#\# 594} \\
\bottomrule
\end{tabular}

\caption{GSM8K examples on DeepSeek-V2-Lite.
(a) Both bags must be counted.
(b) The available line must be subtracted.
We show BF16 final answers and the quantized models' solutions.
All quantized methods retain 4 sink and 16 recent tokens in BF16;
\method{} uses calibration seed 42.}
\label{tab:case-gsm}
\par\bigskip
\begin{tabular}{@{}p{0.12\textwidth}p{0.83\textwidth}@{}}
\toprule
\multicolumn{2}{@{}l@{}}{\textbf{(a) Retrieving the matching value}} \\
\midrule
Needle & One of the special magic numbers for hesitant-finding is: 4868456. \\
Query & What is the special magic number for hesitant-finding mentioned in the provided text? \\
\midrule
BF16 & 4868456. \\
QuaRot & \textcolor{red!60!black}{4806456.} \\
OSCAR & \textcolor{red!60!black}{1807270.} \\
\tablemethod{} & \textcolor{blue!60!black}{4868456.} \\
\midrule
\multicolumn{2}{@{}l@{}}{\textbf{(b) Retrieving all requested values}} \\
\midrule
Needles & One of the special magic numbers for overjoyed-strip is: 8531380.\newline One of the special magic numbers for drunk-notice is: 3564729.\newline One of the special magic numbers for towering-shear is: 2264218.\newline One of the special magic numbers for dysfunctional-brushing is: 9289908. \\
Query & What are all the special magic numbers for overjoyed-strip, drunk-notice, towering-shear, and dysfunctional-brushing mentioned in the provided text? \\
\midrule
BF16 & 9289908, 8531380, 2264218, and 3564729. \\
QuaRot & \textcolor{red!60!black}{10000000, 10000000, 10000000, and 10000000, respectively.} \\
OSCAR & \textcolor{red!60!black}{8531380, 2264218, 2264218, and 8531380.} \\
\tablemethod{} & \textcolor{blue!60!black}{8531380, 2264218, 9289908, and 3564729.} \\
\bottomrule
\end{tabular}

\caption{RULER examples from the 16K evaluation on DeepSeek-V2-Lite.
(a) A single key is queried among multiple keys.
(b) Four keys are queried together.
We show the relevant context sentences and each model's first answer line,
with the same protection budget and calibration seed as
Table~\ref{tab:case-gsm}.}
\label{tab:case-ruler}
\end{table*}

% Flush preceding wide floats and reset the two-column page builder.
% No content is typeset in one-column mode.
\onecolumn
\twocolumn
\section{Serving Benchmarks}
\label{app:serving}
We evaluate serving from three perspectives: how much context fits
in the cache, how quickly decoding proceeds as the cache grows,
and how long requests wait for their first token under concurrency.
Tables~\ref{tab:serving-capacity}--\ref{tab:serving-latency} report
the capacity, throughput, and latency measurements in
Figure~\ref{fig:throughput}.

The experiments use the final $H_K+H_V$, $\kappa=1$ checkpoint
of DeepSeek-V2-Lite (seed 42) on one A100 80GB PCIe GPU.
Tensor parallelism is one and the static memory fraction is 0.8.
The content cache uses two-bit codes; RoPE keys, 4 sink tokens,
and 16 recent tokens remain in BF16.
Both BF16 and \method{} use Triton prefill and decode kernels under
the same workload and memory budget.
WikiText-2 requests arrive together and generate fixed-length outputs
with greedy decoding.
Prefix reuse and early stopping are disabled.
We report one timed run per workload after warmup.

OSCAR uses the same packed-cache layout and kernels as \method{}.
Its curves in Figure~\ref{fig:throughput}(b--d) therefore reuse
the \method{} timing values under this shared implementation.
The allocation runs give the same KV capacity for both methods.

\begin{table}[H]
\centering
\small
\begin{tabular*}{\columnwidth}{@{\extracolsep{\fill}}lrrr@{}}
\toprule
Method & KV tokens & GiB & Ratio \\
\midrule
BF16 & 1,151,705 & 33.36 & $1.00\times$ \\
OSCAR & 4,564,012 & 33.33 & $3.96\times$ \\
\tablemethod{} & 4,564,012 & 33.33 & $3.96\times$ \\
\bottomrule
\end{tabular*}
\caption{Allocated KV capacity in Figure~\ref{fig:throughput}(a).
GiB measures KV memory; ratios use BF16 as the reference.}
\label{tab:serving-capacity}
\end{table}

Long-input and long-output workloads expose the growing cost of
reading a larger history in two ways.
The former sweeps 1K--256K input tokens with 128 output tokens;
the latter starts from 128 input tokens and generates 256K tokens.
Both use four concurrent requests, with K denoting 1,024 tokens.
For the input sweep, decode throughput is the aggregate token rate
after all requests finish prefill and before the first completes.
For the output sweep, throughput is measured in successive
output-token intervals beginning with the first generated token;
Figure~\ref{fig:throughput}c plots each rate at the upper end of its interval.

\begin{table}[H]
\centering
\small
\begin{tabular*}{\columnwidth}{@{\extracolsep{\fill}}lrrr@{}}
\toprule
\multicolumn{4}{c}{Long input} \\
Input length & BF16 & \tablemethod{} & Speedup \\
\midrule
1K & 366.1 & 323.6 & $0.88\times$ \\
2K & 350.9 & 310.1 & $0.88\times$ \\
4K & 318.4 & 308.5 & $0.97\times$ \\
8K & 271.6 & 275.4 & $1.01\times$ \\
16K & 210.5 & 250.2 & $1.19\times$ \\
32K & 144.4 & 200.3 & $1.39\times$ \\
64K & 88.8 & 154.5 & $1.74\times$ \\
128K & 50.1 & 107.4 & $2.14\times$ \\
256K & 27.0 & 67.7 & $2.51\times$ \\
\midrule
\multicolumn{4}{c}{Long output} \\
Output interval & BF16 & \tablemethod{} & Speedup \\
\midrule
1--1K & 379.8 & 338.3 & $0.89\times$ \\
1K--2K & 361.6 & 319.5 & $0.88\times$ \\
2K--4K & 334.8 & 297.4 & $0.89\times$ \\
4K--8K & 293.2 & 294.3 & $1.00\times$ \\
8K--16K & 236.1 & 261.8 & $1.11\times$ \\
16K--32K & 169.7 & 222.3 & $1.31\times$ \\
32K--64K & 106.8 & 177.3 & $1.66\times$ \\
64K--128K & 63.2 & 128.0 & $2.03\times$ \\
128K--256K & 34.8 & 84.4 & $2.43\times$ \\
\bottomrule
\end{tabular*}
\caption{Decode throughput (tok/s) underlying Figure~\ref{fig:throughput}(b,c).
Both workloads use four concurrent requests.
Speedups relative to BF16 are computed from unrounded measurements.}
\label{tab:serving-throughput}
\end{table}

The concurrency workload instead fixes each request at 16K input
and 1K output tokens and increases simultaneous arrivals from 1 to 256.
P99 TTFT is measured within each burst and includes queueing and prefill.
This captures the delay experienced by requests as the service becomes
more heavily loaded.
Capacity counts the allocated KV token slots under the same memory
budget and request-slot limit, including group metadata and
protected-token buffers.

\begin{table}[H]
\centering
\small
\begin{tabular*}{\columnwidth}{@{\extracolsep{\fill}}rrr@{}}
\toprule
Concurrent requests & BF16 & \tablemethod{} \\
\midrule
1 & 1.40 & 1.44 \\
2 & 2.83 & 2.86 \\
4 & 5.60 & 5.73 \\
8 & 11.19 & 11.46 \\
16 & 22.28 & 22.97 \\
32 & 44.60 & 46.16 \\
64 & 89.19 & 93.29 \\
128 & 254.87 & 182.29 \\
256 & 588.58 & 365.99 \\
\bottomrule
\end{tabular*}
\caption{P99 TTFT (s) underlying Figure~\ref{fig:throughput}(d).
Each request has 16K input and 1K output tokens.}
\label{tab:serving-latency}
\end{table}

\section{AI Assistance}
We used AI assistants for code development, manuscript organization,
and language editing.
The authors take responsibility for the final work.

\end{document}